\documentclass[sigconf]{acmart}

\usepackage[ruled]{algorithm2e} 
\usepackage{graphics} 
\usepackage{wrapfig}
\usepackage{subcaption}

\AtBeginDocument{%
  \providecommand\BibTeX{{%
    \normalfont B\kern-0.5em{\scshape i\kern-0.25em b}\kern-0.8em\TeX}}}

\newcommand{\BF}[1]{\ensuremath{\mathbf{#1}}}

\newcommand{\squeezeuppicture}{\vspace{-3mm}} 

\setcopyright{acmcopyright}
\begin{document}

\title{Privacy is What We Care About: Experimental Investigation of Federated Learning on Edge Devices}


\author{Anirban Das}
\affiliation{\institution{Rensselaer Polytechnic Institute\\
Troy, New York, USA}}
\email{dasa2@rpi.edu}
\authornote{Work done as part of research internship in IBM Research, Zurich}

\author{Thomas Brunschwiler}
\affiliation{\institution{IBM Research Zurich\\
R{\"u}schlikon, Switzerland}}
\email{tbr@zurich.ibm.com}


\begin{abstract}

    \textit{Federated Learning} enables training of a general model through edge devices without sending raw data to the cloud. Hence, this approach is attractive for digital health applications, where data is sourced through edge devices and users care about privacy.
    Here, we report on the feasibility to train deep neural networks on the Raspberry Pi4s as edge devices. A CNN, a LSTM and a MLP were successfully trained on the MNIST data-set. Further, federated learning is demonstrated experimentally on IID and non-IID samples in a parametric study, to benchmark the model convergence. The weight updates from the workers are shared with the cloud to train the general model through federated learning. With the CNN and the non-IID samples a test-accuracy of up to 85\% could be achieved within a training time of 2 minutes, while exchanging less than $10$ MB data per device.
    In addition, we discuss federated learning from an use-case standpoint, elaborating on privacy risks and labeling requirements for the application of emotion detection from sound. Based on the experimental findings, we discuss possible research directions to improve model and system performance. Finally, we provide best practices for a practitioner, considering the implementation of federated learning.

\end{abstract}



\copyrightyear{2019} 
\acmYear{2019} 
\acmConference[AIChallengeIoT'19]{First International Workshop on Challenges in Artificial Intelligence and Machine Learning }{November 10--13, 2019}{New York, NY, USA}
\acmBooktitle{First International Workshop on Challenges in Artificial Intelligence and Machine Learning (AIChallengeIoT'19), November 10--13, 2019, New York, NY, USA}
\acmPrice{15.00}
\acmDOI{10.1145/3363347.3363365}
\acmISBN{978-1-4503-7013-4/19/11}

\maketitle

\section{Introduction} \label{sec.introduction}
Recently, the availability of low-cost and performant edge devices has resulted in a trend of pushing intelligent services towards the edge of the network. Moreover, these devices have resulted in an exponential increase of data collected at the network edge~\cite{gartner21billion}. The availability of such smart devices and data  facilitates the deployment of advanced deep learning algorithms in proximity of the consumer, in effect doing edge computing.
In general, large data sets need to be accessible during model training to yield the required accuracy of deep neural networks. Accordingly, recorded data sets from the edge devices are sent to the cloud to perform the model training on the consolidated data. The trained models are then deployed to the edge devices for inference. For personal sensitive data (e.g. audio and video recordings) this approach suffers from user acceptance. 
User expectations and regulations (e.g. GDPR) drive increasing levels of data security and privacy of personal sensitive data, also for machine learning applications \cite{shokri2015privacy}.
Thus, we explore a distributed machine learning approach named \textit{Federated Learning} to respond to the trend. Federated learning allows training models with data-sets stored on individual clients. The locally optimized model weights are then aggregated centrally to train a general global model. 
By design, some of the privacy concerns can be alleviated as the raw data never leaves the premises of the patient or hospital. Moreover, we can add additional layers of e.g. secure multi-party communication, homomorphic encryption or differential privacy \cite{abadi2016dp, yang2019federated} to increase the level of security further.

\section{Why federated learning?} \label{sec.motivation}
        \textbf{Federated Learning:} The term \textit{Federated Learning} was first proposed by \cite{mcmahan2016communication}, with the algorithm \textit{FedAvg}. Here, training happens in multiple \textit{rounds} of communication. The central server first sends the current global model to the clients/workers (we use the term client and worker interchangeably).
        The set of clients then perform local optimization and communicate the learned model weights to the central server. The central server learns a global shared model by methodically aggregating the updates from the clients. Preferably, the training on the clients is performed over several batches to minimize network traffic.
        
        In many cases, data owners have constraints on directly sharing their raw data for model training on the cloud due to either privacy, security or legal constraints. However, they could benefit from a cooperative learning process to train the global model. Federated learning can be appropriate in this context. Only the weights of the model representing a statistical summary of many raw data samples are sent to the cloud, resulting in an inherently privacy preserving approach. 
        Further, for federated supervised learning to work, we need the presence of labelled data at the edge. Labels can be obtained from various avenues such as the user self reports or from reports by non-colluding experts during training. In case of unsupervised learning, however, no labels are required. 

\textbf{Data Distribution}: The availability of independent and identical distributed (IID) and non-IID data for a worker depends on the use-case. In health-care applications, a single patient, with a rather personalized data distribution is the source of data per edge device, resulting in non-IID training. However, in scenarios of predictive maintenance of machinery, with identical device types as data source, the data can be assumed to be IID. Thus, we explore the importance of data distribution types an the training performance.

\begin{figure}[t]
\centering
	\begin{subfigure}{0.49\linewidth}
      \centering
      \includegraphics[width=\linewidth]{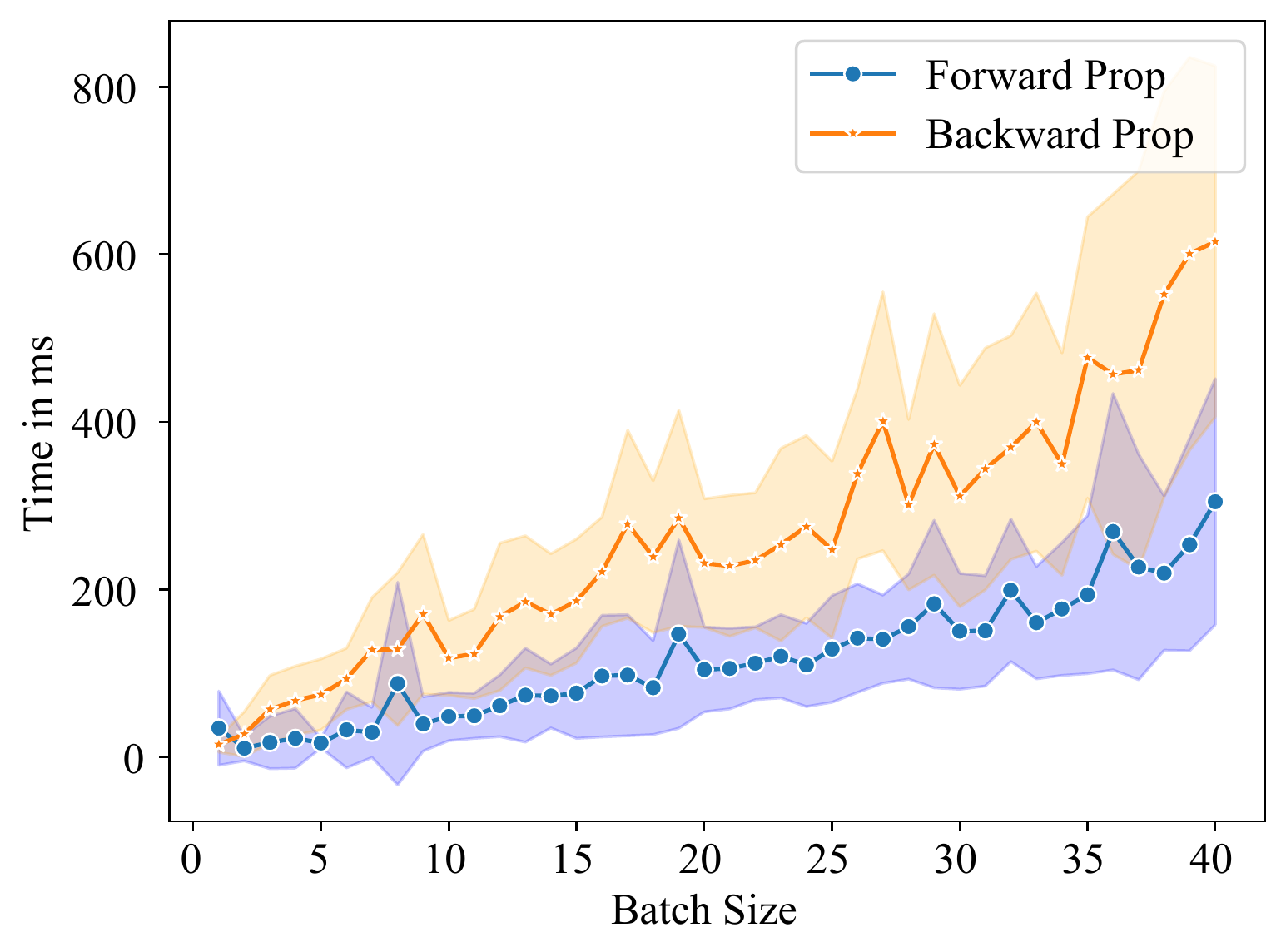}
      \caption{CNN}
      \label{fig.cnn}\squeezeuppicture
   \end{subfigure}~~~~%
	\begin{subfigure}{0.49\linewidth}     
      \centering
      \includegraphics[width=\linewidth]{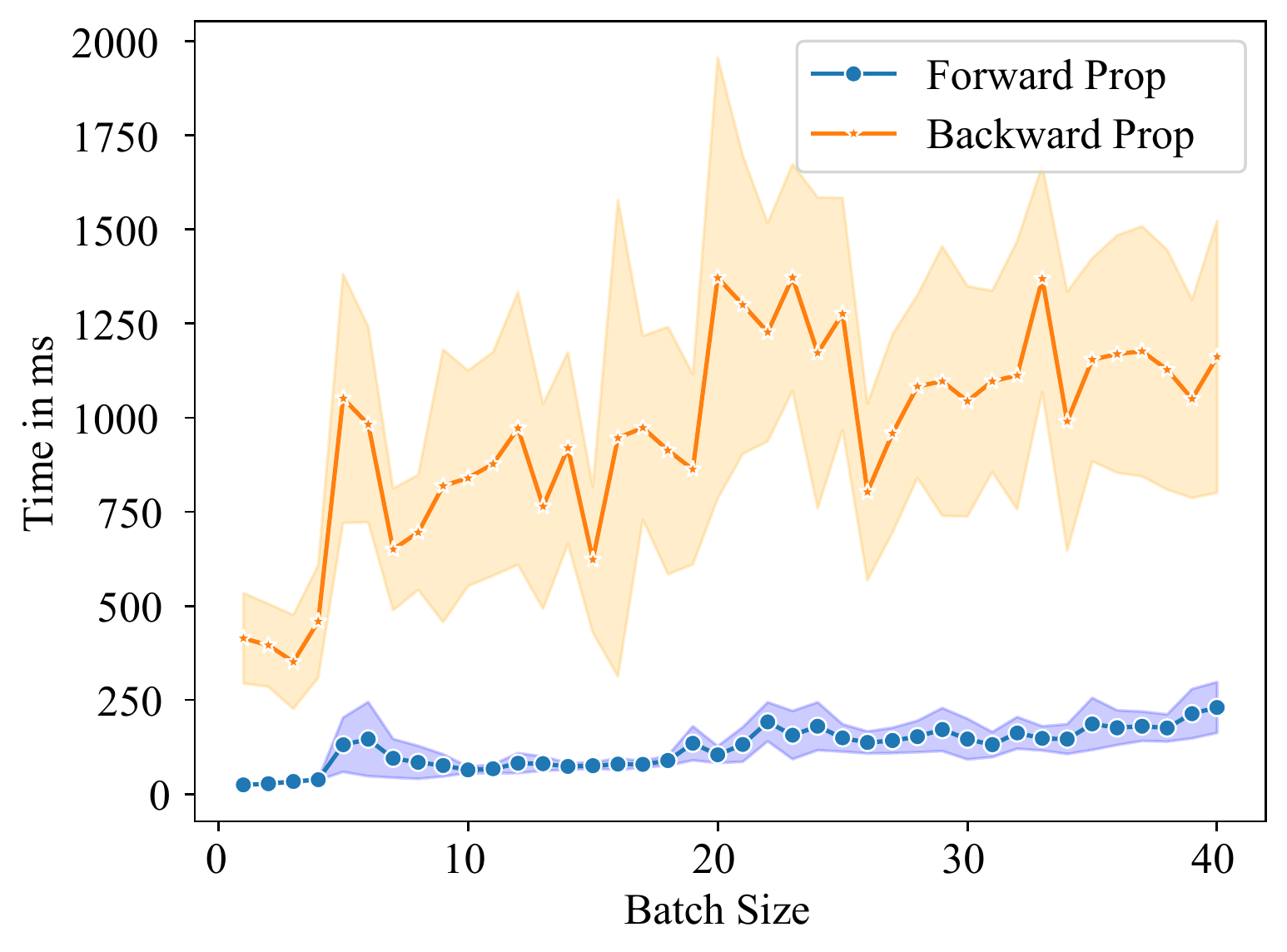}
      \caption{LSTM}
      \label{fig.lstm}\squeezeuppicture
   \end{subfigure}~~~~%
   \caption{Forward and backward propagation latency versus batch size}
   \label{fig.feasibility}
   \squeezeuppicture
\end{figure}
\section{Learning at the edge} \label{sec.sysarch}
\begin{wrapfigure}{r}{0.47\linewidth}
  \begin{center}
    \includegraphics[width=\linewidth]{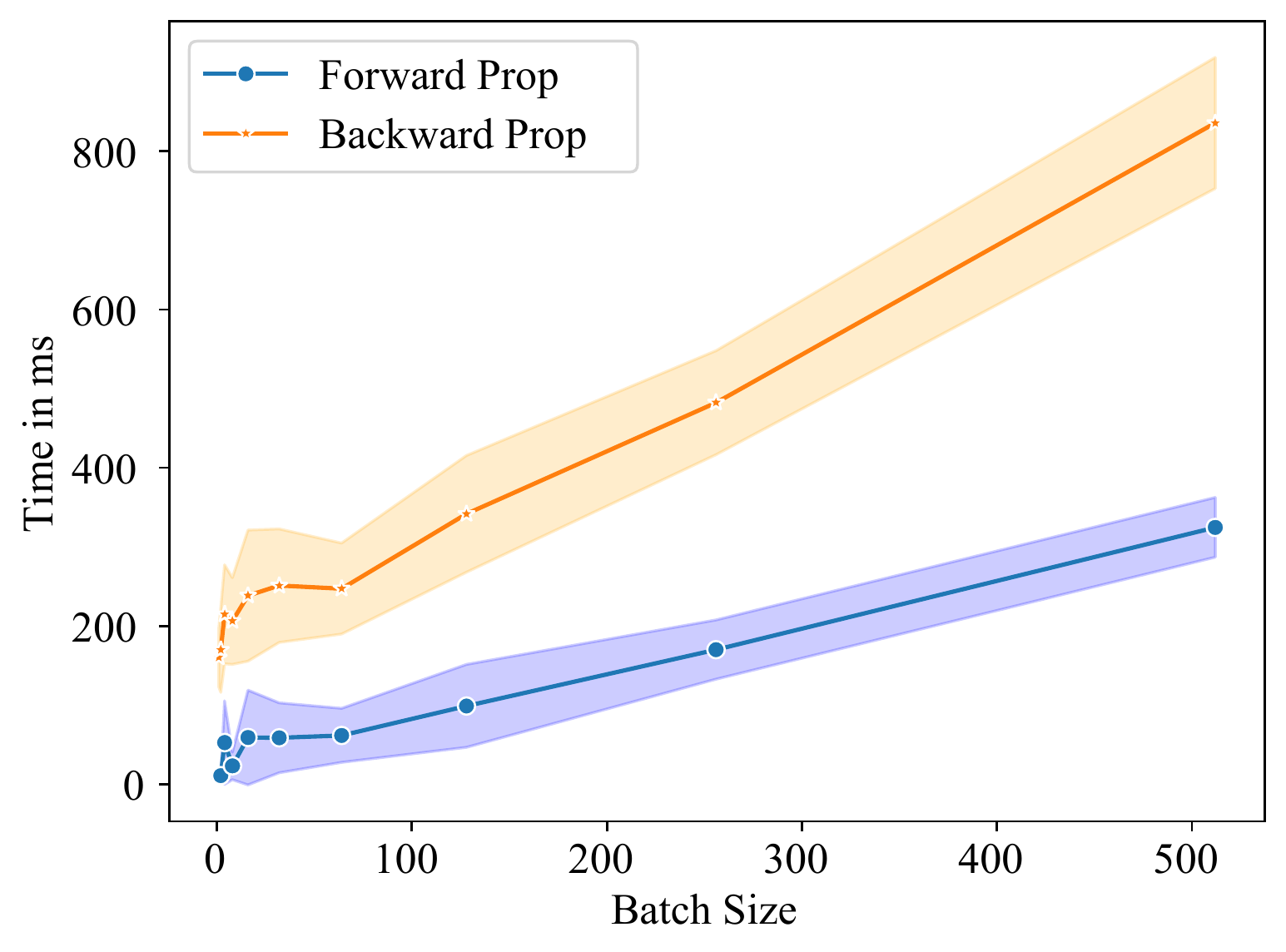}
  \end{center}
  \caption{Forward and back propagation latency versus batch size in MLP}
     \squeezeuppicture
    \label{fig.feasibilityMLP}
\end{wrapfigure}

For federated learning, execution of substantial machine learning (training) workloads needs to be feasible on resource constrained edge devices.
In order to explore the feasibility of training, we consider the newly introduced Raspberry Pi 4 devices (with 1.5 GHz 4 core processor and 4GB RAM) as the edge device. Further, we assume the edge devices are connected to steady energy sources. For our experiments, we do not deploy sensors, but store raw data as files on the workers (i.e. image files from the MNIST data-set). 
We consider 3 network models for the feasibility study: 

\begin{footnotesize}
\begin{enumerate}
    \item CNN: Conv2d(16) $\rightarrow$ Conv2d(32) $\rightarrow$ Maxpool(2) $\rightarrow$ FC(10) ; \textit{(47K params)}
    \item LSTM: LSTM(128) $\rightarrow$ LSTM(64) $\rightarrow$ FC(8) ; \textit{(641K params)}
    \item MLP: FC(512) $\rightarrow$ FC(256) $\rightarrow$ FC(10) ; \textit{(1.7M params)}
\end{enumerate}
\end{footnotesize}
In Fig.~\ref{fig.feasibility} and Fig.~\ref{fig.feasibilityMLP} we show the latency for forward and backward propagation during training vs. the increase in batch size. The input to the CNN and the MLP are random matrices of size ($1\times28\times28$) and ($1\times 3*32*32$) respectively to simulate the dimensionality of common data-sets of MNIST and CIFAR. The input to the LSTM is ($8\times1024$) with 8 timesteps and 1024 dimensions to simulate an input embedding from SoundNet, which we aim to use for Speech Emotion Recognition (discussed in Sec.~\ref{sec.use_case}). We measure the actual CPU time spent in forward and backward propagation using PyTorch's `autograd profiler' and present the mean results over 50 runs per batch size.\\
\textbf{Results and Discussion:}
We observe that for all 3 cases, the back propagation is substantially slower than the forward pass. For the CNN we observe a linear increase with the same factor in CPU time vs. batch size. The latency for processing a batch size of 32 is $<600$ms. Further, we observe that the variance of both latencies increase with batch size in case of the CNN. For the LSTM, the forward propagation scales linearly with batch size. The difference between forward and back propagation is much larger and back propagation latency has a weak linear trend. In our experiments, we went up to moderate batch-size of 40, yet the Raspberry Pi4 was still able to support the load with an average delay of $<1.4~s$. In older Raspberry Pi3 we started getting `seg-faults` from batch size 8 with the same model prompting training impossible. The fact that we can easily go over batch size 40 in Raspberry Pi4 without memory error is already a huge improvement. 
For the MLP, we observe that the forward and backward propagation time for batch sizes up-to 32 does not vary much and stays below $\approx 250 ms$ for back propagation, and $\approx 60 ms$ for forward propagation with some jitters. For larger batch size of up to 512 we see both the latencies go up linearly, with low variance. This suggests that up-to a certain problem size, the bottleneck is most likely in the framework and not in the compute performance of the hardware. Also, the iterations in the MLP are much faster than the CNN or RNN inspite of higher number of parameters. We note that compute time not only depends on the number of parameters or operations, but also on other factors such as framework implementation, memory access patterns, caching effects, special mathematical operation overheads, compiler optimization etc. If there are time constraints for training, MLP network could therefore be a good choice given it performs well in the use-case.

In conclusion, with the new Raspberry Pi4 we can train small to moderate networks with reasonable batch sizes. Moreover, the latency to train each iteration over such batch sizes is also very low and can be done in several hundred milliseconds to a few seconds. The hardware improvement over the previous generation Raspberry Pi3 has mainly contributed to its usefulness for \textit{training} deep networks.

\begin{figure*}[t]
\centering
    \begin{subfigure}{0.27\linewidth}     
      \centering
      \includegraphics[width=\linewidth]{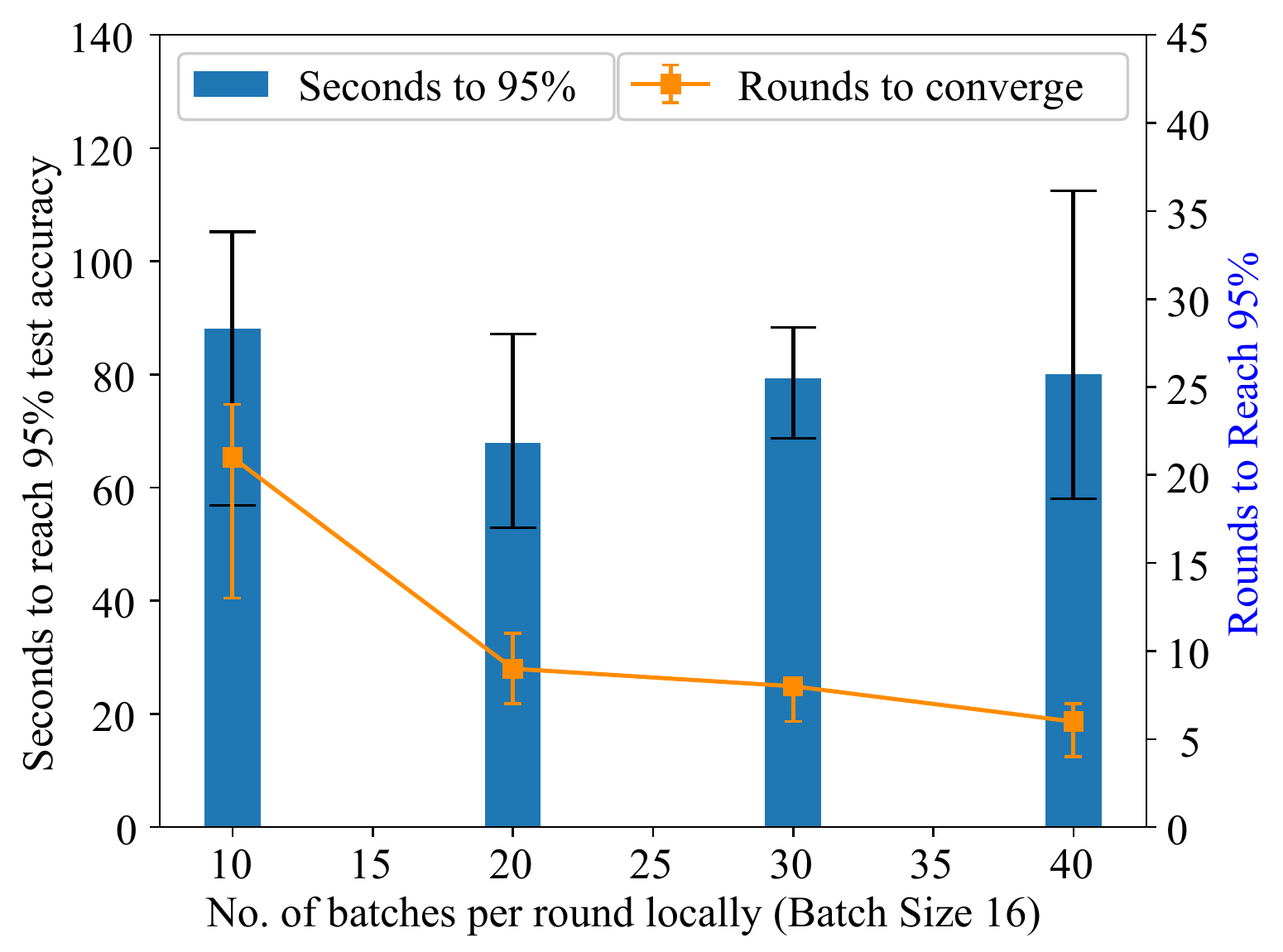}
      \caption{IID}
      \label{subfig.iid_time_bar}\squeezeuppicture
   \end{subfigure}~~~~%
	\begin{subfigure}{0.27\linewidth}
      \centering
      \includegraphics[width=\linewidth]{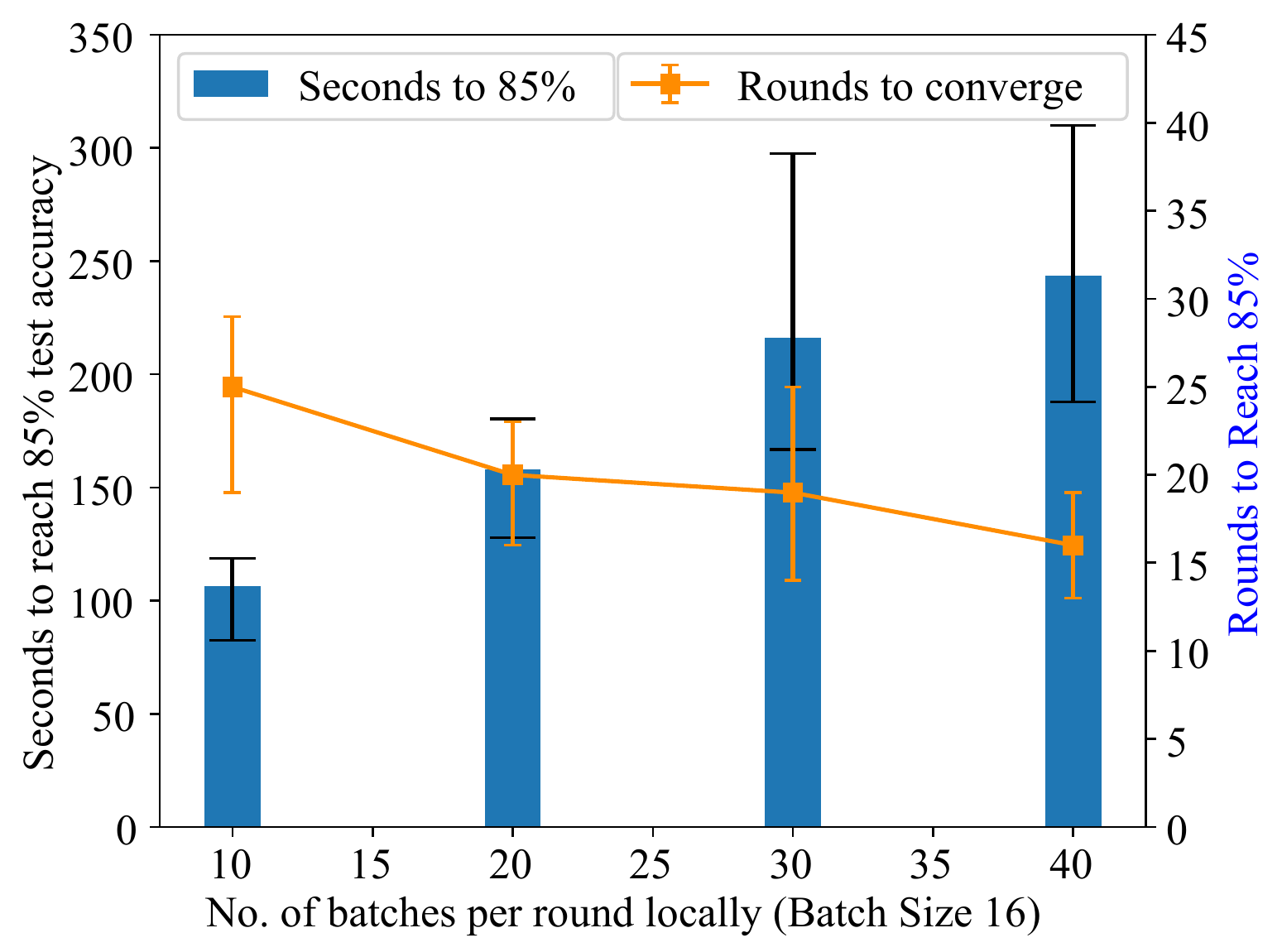}
      \caption{Non-IID}
      \label{subfig.noniid_time_bar}\squeezeuppicture
   \end{subfigure}~~~~%
   \begin{subfigure}{0.27\linewidth}
      \centering
      \includegraphics[width=\linewidth]{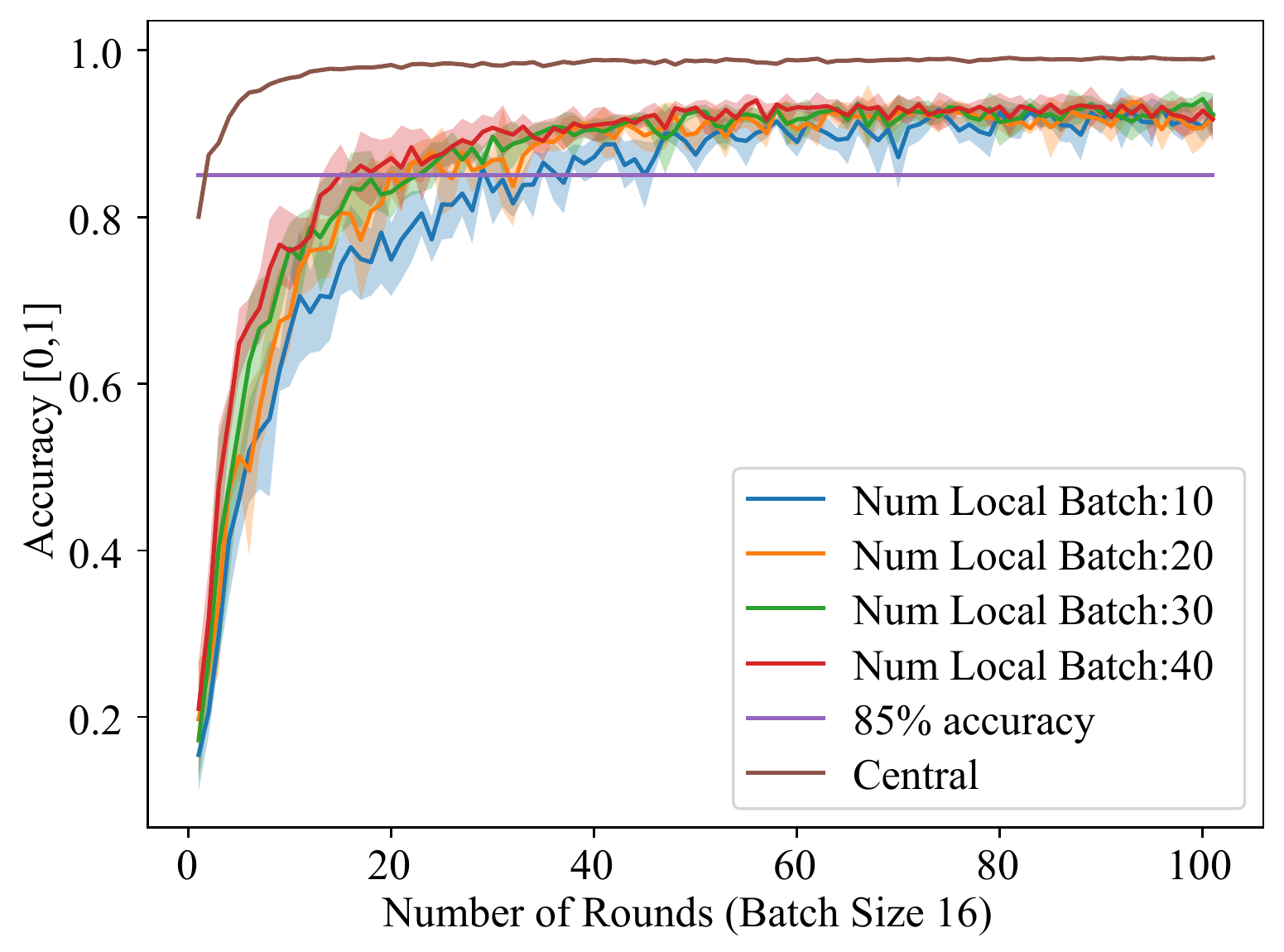}
      \caption{Non-IID Convergence vs. \BF{E}}
      \label{subfig.noniid_conv}\squeezeuppicture
   \end{subfigure}
   \caption{Convergence vs. number of local batches per round}
   \label{fig.convergence}
   \squeezeuppicture
\end{figure*}
\section{Federated Learning Performance}
\label{sec.experiments}
In this section we study the performance of federated learning on resource constrained edge devices.
We use an internally developed federated learning framework for our experiments. Experiments with PySyft\footnote{https://github.com/OpenMined/PySyft} and comparative bench-marking is a future goal. 
Hardware-wise, we created a test-bed consisting of 5 Raspberry Pi 4s. Each device is connected to the central server, a Macbook Pro 
 via a 300 Mbits wireless router. We have used PyTorch v1.1.0 and all code developed in Python v3.7.
Experiments are performed using the \textit{FedAvg} algorithm. 
In the experiment
we used all 5 clients ($K=5$) for each round. We define \textit{round} as one iteration of the whole \textit{FedAvg} process. 
Further, iterating through all local data points per round is difficult for resource constrained workers. For ease of performance characterization, we re-define \BF{E} to denote the number of local batches of fixed size each worker trains in each round  and \BF{B} as the batch size. The magnitude of hyperparameters \BF{E} and \BF{B}, determines the amount of local computation and thereby training performance.
We use the standard image classification task on the MNIST data-set as the federated learning task, to keep our study comparable to the literature \cite{abadi2016dp,mcmahan2016communication, caldas2018leaf, geyer2017differentially}. 

        
        

 In the experiment, we studied the compute vs. communication latency of individual workers and the convergence behaviour of federated learning using the CNN from Sec.~\ref{sec.sysarch}. 
Furthermore, \BF{E}, the number of batches on which training is executed on a client per round was varied, while keeping the batch size fixed at \BF{B}=16. It should be noticed, for IID distribution, the data is distributed randomly among 5 devices, for the non-IID distribution the 5 workers each contains non overlapping data points corresponding to 2 digits (ie. [0,1], [2,3], ...). The accuracy test is done on the central server device on the test set. \cite{shiqiang}
 has shown that synchronous gradient updates converge faster than synchronous federated learning. Hence we use synchronous gradient updates. 

 For both, the IID and the non-IID case, we run 5 experiments each with $\BF{E} \in \{10, 20, 30, 40\}$. For each experiment we record the \textit{round} and corresponding time when the system \textit{first} achieves 95\% and 85\% test accuracy for the IID and for the non-IID data-sets, respectively. In Fig.~\ref{subfig.iid_time_bar},\ref{subfig.noniid_time_bar} the bars represent the average wall clock time, the error bars represent the \textit{min} and \textit{max} time observed among 5 runs for each value of \BF{E} (measured at the central server). We also plot the average round (quantized to closest integer) and min and max rounds needed in the 5 runs for each value of \BF{E}. Further, the wall clock time of each device spent on compute, communication or on idle time is recorded, besides the data transmission between the workers and the server using \texttt{pyshark}\footnote{https://github.com/KimiNewt/pyshark}. \\
\textbf{Results and Discussion:}
We observe that training on IID data converges to 95\% test-accuracy in as low as 6 rounds of federated averaging at \BF{E}=40. As expected, with increasing \BF{E} the number of rounds needed for convergence is dropping rapidly. But once the local computation reaches a certain level \BF{E}$=$30 or \BF{E}$=$40, the local models generalize themselves enough, so further training does not improve the model accuracy. 
 \begin{figure}[t]
\centering
\includegraphics[width=0.7\linewidth]{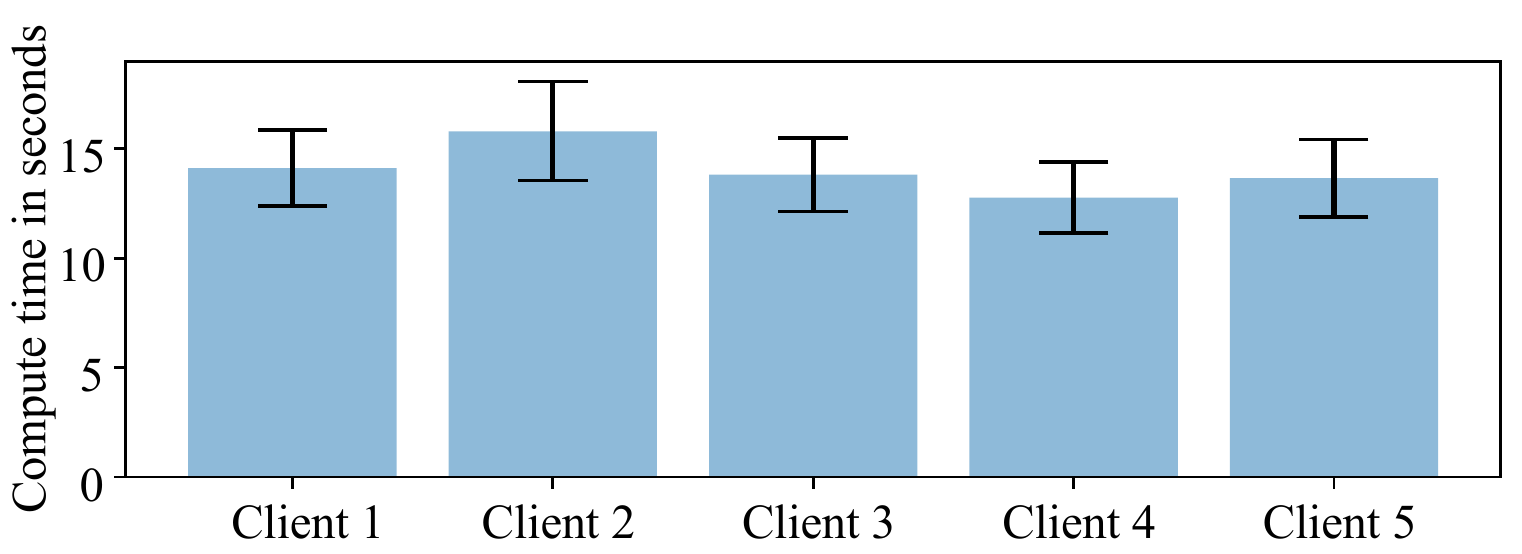}
\squeezeuppicture
    \caption{Avg. compute time at all clients}
        \label{subfig.compute_time_all}
\squeezeuppicture
\end{figure}
For the non-IID case, we observe that the number of rounds to reach 85\% accuracy does only decrease moderately with increalocal training per round i.e. increase of \BF{E}. In the non-IID case the individual models on clients can diverge substantially by increasing the number of times the local model is updated before the global averaging. Hence, the system needs considerable more time to converge when we increase \BF{E}. In general, the actual wall clock time for non-IID is larger than that in IID. However, we still observe good performance (even with limited local computation) with an accuracy of 85\% within 2 minutes considering \BF{E}=10 and in 25 rounds. In Fig.\ref{subfig.noniid_conv} we observe good convergence, where we plot the accuracy and its variance per round averaged over 5 runs.

It has to be mentioned, that the variability in performance of the Raspberry Pis is large. 
We pick an example run from the non-IID experiments (\BF{E}=40, \BF{B}=16) to demonstrate this variability for a fixed computation size, however, with varying data points in batches. The average compute time across 100 rounds is depicted in Fig.~\ref{subfig.compute_time_all} for individual clients. 
On average, worker 2 seems to be the straggler in this run. Since we have synchronous rounds of communication, the server waits till it receives model updates from each of the devices. Overall, the maximum average compute time per round of training in a client here is $\approx$ 15s. 
On investigating the different components of latency of the slowest worker 2 we further observed that the average compute time on a device is an order of magnitude larger than the time the device spends to send or receive the model weights, rendering the former the bottleneck in the cycle.

 We find that each worker exchanges $\approx 400~KB$ back and forth with the server in each round. This is once for downloading the global model weights and sending the updated weights for each round, respectively. For test set convergence to 85\% in the non-IID case, the total data transmission between the server and each worker is $\approx 10~MB$ for \BF{E}=10 up-to round 25, which is quite low, but has potential to result in substantial traffic with more devices. 


\section{Usecase and Recommendations}\label{sec.discussion}
\subsection{Possible attack and data loss scenarios}
When data is sent to the cloud for training, adversarial attacks on the central server may compromise information of all clients, resulting in a massive security breach scenario, affecting millions of users. Vanilla federated learning prevents leakage of raw data from the cloud, by transferring only statistical aggregates, i.e. the model weights to a central server. Thus, data would only leak from individual devices, compromising only one or few users. However, addition of techniques such as differential privacy could improve the robustness against adversarial attacks even further. 


\subsection{Usecase study}\label{sec.use_case}

One interesting federated learning use-case is emotion detection from audio signals. This could be useful 
in elderly care homes for detecting depression or general mood of the elderly to take precautionary health measures and their improving quality-of-life. With permission of the elderly, a small edge device can be installed in the room to intermittently record utterance level audio from the person.  
Now, due to the sensitive nature of the information which was captured, in accordance with the GDPR and other privacy laws, it is imperative that the processing and training is taken place on premises. 
For labelling, either the elderly could be asked to fill out a self-report or the concerned nurse or doctor could perform the labeling of the emotion of the recorded audio clips each day.
Here, we could make use of federated learning, where each Raspberry Pi in the room of each elderly acts as a worker device. The first tier of privacy concern is alleviated as the raw data is not transmitted to the cloud. Further, if weights are intercepted, it can be very hard to reconstruct an audio signal from compromised model weights and thus, the motivation to do so depends on the cost benefit for the adversary. Additionally, differential privacy techniques such as gradient clipping and addition of Laplace or Gaussian noise to the weights transmitted as done by authors in \cite{geyer2017differentially} could be considered. 

\subsection{Challenges and opportunities} \label{subsec.challenges}
\textbf{Communication Structure}: Traditionally, federated learning is hierarchical with edge nodes and one central `cloud' server. We can envision using fog computing for hierarchical gradient aggregation and have aggregate models corresponding to different geo-spatial scale at different hierarchies instead of a single central server. However, ensuring security and privacy will be challenging.\\
\textbf{Hardware Limitations}: Sometimes the learning task can be very hard and resource hungry. A good example is emotion detection from audio, which needs advanced deep networks such as FCN, TCN etc., and may be infeasible to run on resource constrained edge devices. Here, we could train a general model on the cloud using the full network architecture and later fine tune it by only training the last few layers using federated learning at the edge. \\
\textbf{Stragglers}: For synchronous gradient updates, the straggler nodes contribute substantially to the overall latency. One solution can be to accept \textit{approximate} gradient updates from workers as proposed by authors in \cite{smith2017federated}. Another solution can be to piggyback the compute and resource utilization with client updates.
Based on this, a predictive performance model for the workers can be formed and dynamically adjust to the amount of local computation by controlling \BF{E} and batch size. \\
\textbf{Personalization}: The voice pattern and emotional state of each person is different, thus, the generated data will be highly non-IID. Hence, it may be effective to maintain a personalized model locally, alongside the global model. 
For this purpose, we can perform transfer learning at the edge by transferring the frozen global model to the edge and retrain a fully connected layer or an SVM on local data using the global model as feature extractor. Federated multitask learning proposed in \cite{smith2017federated} could also be used for personalization.
\textbf{Framework Limitations}: So far, there is a dearth of frameworks to seamlessly deploy federated learning workloads. PySyft and Tensorflow Federated are the prominent leaders, but their deployability and scalability are limited. A possible way may be to use Kubernetes cluster as the central hub where nodes may join as workers. It could also be interesting to integrate federated learning with edge computing frameworks like AWS Greengrass, Azure IoT Edge, KubeEdge etc. to deploy worker logic.\\
\textbf{Benchmarking}: As more frameworks emerge, it is important to evaluate those systems with standardized workloads and representative use cases on real testbeds to understand trade-offs in performance and decide future bottlenecks. So far, a large part of the literature has internal implementations of federated learning framework due to reasons mentioned above. \cite{caldas2018leaf} provides an excellent benchmark suite, but we need more such studies to tackle the breadth of application requirements.

\section{Conclusions} We briefly studied the feasibility of federated learning on resource constrained edge devices. We demonstrated the capability of current  devices to train deep neural networks and presented the performance numbers on MNIST data-set trainings. For practitioners considering health-care applications with vital signs, audio or video data (data with limited damage) we recommend a) RISK: federated learning should be applied to limit data leaks to individual clients (edge devices) rather than millions of data samples from the cloud, b) TRAINING: for model training speed, consider low number of batch per round in case of non-IID data, but keep a minimum batch size to ensure enough learning from the data. c) DATA TRANSFER: In the IID case, the number of batch per round should be increased to minimize data transfer and according cost (especially if communication over channels inferior to WiFi are used). Federated learning on edge devices is an interesting new field and we hope to see more exciting research in this area.


\begin{acks}
We want to thank Supriyo Chakraborty, Dean Steuer, Shalisha Witherspoon from IBM Research (Yorktown) for providing us support and insights in using the federated learning framework.
\end{acks}

\bibliographystyle{ACM-Reference-Format}
\bibliography{acmart}
\end{document}